\documentclass[preprint,12pt]{elsarticle}

\usepackage{amssymb}
\usepackage{amsmath,amsfonts}
\usepackage{graphicx}
\usepackage{multirow}
\usepackage{booktabs}
\usepackage{makecell}
\usepackage{caption}
\usepackage{subfig} 
\usepackage{rotating}
\usepackage{threeparttable}

\journal{ArXiv}
\date{}

\begin{document}

\begin{frontmatter}



\title{STG4Traffic: A Survey and Benchmark of Spatial-  Temporal Graph Neural Networks for Traffic Prediction}





\nonumnote{}

\author[1]{Xunlian Luo}
\ead{xlluo@stu.suda.edu.cn} 

\author[2]{Chunjiang Zhu}
\ead{chunjiang.zhu@uncg.edu}

\author[1]{Detian Zhang \corref{a}}
\ead{detian@suda.edu.cn}

\author[3]{Qing Li}
\ead{qing-prof.li@polyu.edu.hk} 

\address[1]{
Institute of Artificial Intelligence, Department of Computer
Science and Technology, Soochow University, Su Zhou, China.}
\address[2]{Department of Computer Science, UNC Greensboro, NC, USA.}
\address[3]{Department of Computing, The Hong Kong Polytechnic University, Hong Kong, China.}

\cortext[a]{Corresponding author.}

\begin{abstract}
Traffic prediction has been an active research topic in the domain of spatial-temporal data mining. Accurate real-time traffic prediction is essential to improve the safety, stability, and versatility of smart city systems, i.e., traffic control and optimal routing. The complex and highly dynamic spatial-temporal dependencies make effective predictions still face many challenges. Recent studies have shown that spatial-temporal graph neural networks exhibit great potential applied to traffic prediction, which combines sequential models with graph convolutional networks to jointly model temporal and spatial correlations. However, a survey study of graph learning, spatial-temporal graph models for traffic, as well as a fair comparison of baseline models are pending and unavoidable issues. In this paper, we first provide a systematic review of graph learning strategies and commonly used graph convolution algorithms. Then we conduct a comprehensive analysis of the strengths and weaknesses of recently proposed spatial-temporal graph network models. Furthermore, we build a study called STG4Traffic using the deep learning framework PyTorch to establish a standardized and scalable benchmark on two types of traffic datasets. We can evaluate their performance by personalizing the model settings with uniform metrics. Finally, we point out some problems in the current study and discuss future directions. Source codes are available at https://github.com/trainingl/STG4Traffic.
\end{abstract}

\begin{keyword}
Traffic prediction  \sep
Urban computing  \sep
Graph neural networks  \sep
Spatial-temporal data mining  \sep
Benchmark
\end{keyword}

\end{frontmatter}


\section{Introduction}
With the rapid development of the Internet of Things (IoT) and urban computing, the massive deployment of sensors provides a reliable source of data for intelligent transportation systems \cite{DBLP:journals/percom/NagyS18, DBLP:journals/tits/YeZYX22}. To alleviate the pressure of the growing population and vehicles in urbanization, research on data-driven traffic systems has become a hot topic in academia and industry. Traffic prediction, as a fundamental-level task of Intelligent Transportation System (ITS), supports a large number of upper-layer applications on the traffic scene, such as congestion warning, route planning and location services \cite{DBLP:journals/tits/YinWWSQY22}. Traffic forecasting is achieved through the statistics, analysis and summary of historical traffic data to realize the judgment of future flow trends \cite{tedjopurnomo2020survey}. In traffic management and control systems, accurate traffic forecasting can help city managers perceive the health of the traffic road network in real-time, adopt timely solutions to optimize the traffic flow, and thus improve road traffic efficiency. In addition, online maps (e.g., Google Maps, Baidu Maps) can improve the quality of urban services by planning routes in advance for 
travelers and shortening travel time.

\begin{figure}[!t]
\centering
\subfloat{\includegraphics[width=2.9in]{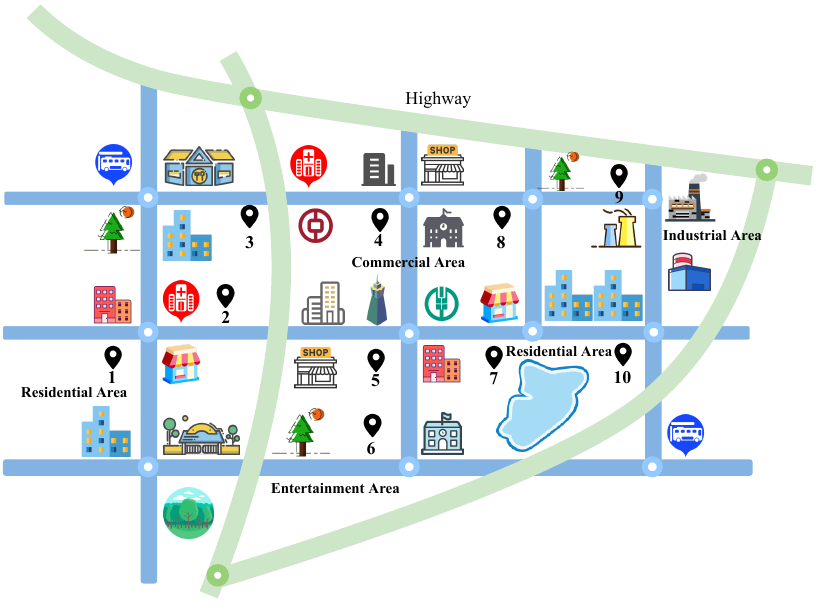}%
\label{fig:spatial}}
\hfill
\subfloat{\includegraphics[width=2.4in]{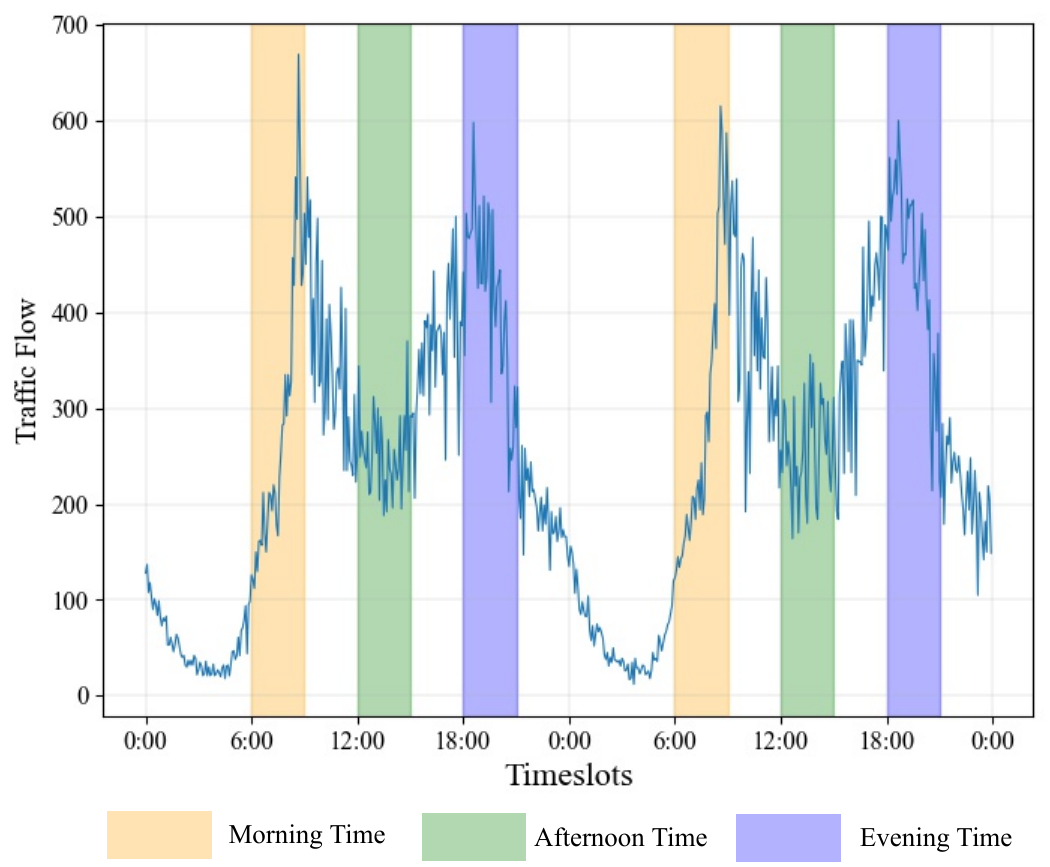}%
\label{fig:temporal}}
\caption{An example of different spatial-temporal correlations among routes. As shown in the left figure, spatial correlation is not only related to the distance, but also to the POI properties of the nodes, and the region similarity. The right figure presents the traffic trend of a node in two days, and the traffic pattern is basically the same for two consecutive days. Meanwhile, within one day, the traffic flows in the morning, afternoon and evening show significant variability.}
\label{fig:correlations}
\end{figure}

Traffic prediction exhibits typical spatial-temporal correlations. As shown in Fig. \ref{fig:correlations}, traffic variation exhibits intricate and multifaceted patterns of spatial and temporal interdependence. In terms of time, traffic volumes (such as, flow, speed, and demand) are affected by the living routine of urban residents and show significant periodicity, e.g., weekdays morning and evening peaks and weekend/holiday aggregated traffic flow \cite{DBLP:journals/isci/WangLLLZL23}. The traffic at an observation point is closely related to the traffic state in the periods before and after, showing certain closeness and trend. In addition to the temporal properties, the intuitive traffic volume changes are also reflected in the information transmission between nodes in the traffic network. Unlike temporal correlations, the potential spatial relationships are diverse \cite{DBLP:conf/ijcai/ShaoJWKXMZZS22, DBLP:conf/aaai/GengLWZYYL19}, as illustrated in Fig. \ref{fig:spatial}, where node 7 and node 10 are connected on the same road with essentially the same traffic patterns. Both node 7 and node 1 belong to the residential area and have significant semantic similarities despite their physical distance. Node 7 and node 4 have the same location function (Same POI, i.e., School, Bank), and even though they are not directly connected, they have the similar spatial pattern \cite{DBLP:journals/corr/abs-2104-14917}. These complex and changing spatial-temporal properties make accurate traffic forecasting still challenging.

Extensive research has been devoted to address the challenge of modeling spatial-temporal data. The earliest statistical models (e.g., VAR \cite{lu2016integrating}, ARIMA \cite{kumar2015short}) are widely used for time series forecasting because of their simplicity and interpretability. However, these designs with restricted parameters are difficult to accomplish complex pattern recognition and the data cannot satisfy the assumption of stationary. Although machine learning methods (e.g., SVR \cite{wu2004travel}, FNN \cite{park1999forecasting}) are often good at non-linear representation, the performance of the models heavily depends on feature engineering and expert experience. Data-driven deep learning techniques, especially temporal convolution \cite{DBLP:journals/corr/abs-1803-01271}, recurrent neural network (and its variants LSTM \cite{hochreiter1997long}, GRU \cite{DBLP:conf/emnlp/ChoMGBBSB14}) and Transformer \cite{DBLP:conf/nips/VaswaniSPUJGKP17}, have made breakthroughs in sequence tasks. However, they treat traffic data as independent signal streams \cite{DBLP:conf/www/Wang0WJWTJY20}, ignoring or barely exploiting the spatial dependence information.

One attempt is to divide the spatial region into same-size grids. ST-ResNet \cite{DBLP:conf/aaai/ZhangZQ17} implicitly represents the correlations between variables in fixed-size convolution kernels by using deep convolutional networks. However, due to the irregularity of roads, topological information inside the traffic network is inevitably missing grid modeling. Inspired by graph neural networks modeling topology graphs, STGCN \cite{DBLP:conf/ijcai/YuYZ18} first proposed to stack gated temporal convolution with graph convolution into Spatial Temporal Blocks to achieve spatial-temporal prediction. This practice demonstrates that embedding prior knowledge of the traffic graph is beneficial to improve the model's predictive performance greatly. In later model design, extensive research efforts have integrated graph neural networks into sequential models to jointly model the potential temporal and spatial dependencies of traffic data \cite{DBLP:conf/cikm/ShaoZ00X22}, and have achieved state-of-the-art performance. Some models such as DCRNN \cite{DBLP:conf/iclr/LiYS018} integrates diffusion convolution into GRU to propose a multi-step prediction architecture that can capture bidirectional random walking graph signals. MTGNN \cite{DBLP:conf/kdd/WuPL0CZ20} designed a structure that combines adaptive graph learning with dilation convolution to capture spatial-temporal correlation. 
GMAN \cite{DBLP:conf/aaai/ZhengFW020} uses spatial-temporal attention fusion to expand the perceptual domain of information but reduce the loss of long-term prediction.

Over the years, although a considerable number of spatial-temporal graph neural network models have been proposed for traffic prediction, the existing literature lacks comprehensive surveys specifically focusing on graph learning and graph computing. While some studies \cite{DBLP:journals/dase/YuanL21, DBLP:journals/eswa/JiangL22} attempt to comprehensively and meticulously organize all the datasets and methods for traffic tasks, they are too broad and fall short in providing precise insight into the core issues and methods of spatial-temporal prediction modeling. Unfortunately, the absence of a fair and standardized benchmark is a significant drawback in the field. Existing benchmarks such as \cite{DBLP:conf/adma/QuachYVNNJ22, DBLP:conf/cikm/JiangYWWDLCDSS21, DBLP:journals/corr/abs-2104-14917} suffer from either irregular experimental settings and limited scalability or exhibit inconsistent results compared to the original papers. Furthermore, the evaluation of these diverse models remains confusing and lacks proper organization.

 To address the above problems, a focused, well-understood, and inslightful survey will be of significance to the development of traffic prediction. In this paper, we  highlight graph structure designs and graph computation methods used in spatial modeling, followed by a concrete survey of spatial-temporal graph neural networks. Then we propose a standardized and easily extensible benchmark to evaluate the performance of the different models. Lastly, we conclude with a prospective analysis of the difficulties and challenges in this study, and possible solutions to resolve them. 
 
 In summary, the main contributions of our paper are provided as follows:
\begin{itemize}
\item Firstly, we provide a comprehensive overview of graph structure design and graph computation methods employed in spatial-temporal graph modeling. This aspect has received less attention in previous surveys, making our discussion particularly valuable.
\item Secondly, we conduct an in-depth survey of spatial-temporal graph neural networks used for traffic prediction. We categorize these methods into three groups based on their temporal characteristics: CNN-Based, RNN-Based, and Attention-Based. Furthermore, we analyze the technical details and limitations of each specific model.
\item Thirdly, we introduce a benchmark named \textbf{STG4Traffic}, which facilitates a comprehensive evaluation of approximately 18 models on traffic speed datasets (METR-LA, PEMS-BAY) and traffic flow datasets (PEMSD4, PEMSD8). Our benchmark yields results that closely align with those reported in the original papers. It not only offers a common data access interface but also provides a unified model training pipeline  for future studies in model design.
\item Lastly, we outline the challenges encountered in traffic modeling from the perspective of data quality, research perspectives, and migration methods. We aim to provide feasible approaches to overcoming the difficulties faced in this field.
\end{itemize}

\section{Problem Statement}
The traffic network can be abstracted as a graph $\mathcal G = (V, E, A)$. $V$ is the set of $N=|V|$ nodes, which represent different observation locations (e.g., traffic sensors, roadway monitoring stations) distributed in the road network.  $E$ is the set of edges and $A \in \mathbb{R}^{N \times N}$ denotes the adjacency matrix depicting the relations between nodes, where each element represents the quantification of proximity from a certain insight, such as road connectivity, distance proximity, POI similarity, etc.

The traffic data observed on $\mathcal G$ at time $t$ is denoted as the graph signal $X_t \in \mathbb{R}^{N\times D}$, and the signal of the $i$-th node is denoted as $X_t^{(i)} \in \mathbb{R}^D$, where $D$ denotes the feature dimension of the node (e.g., traffic flow, speed, and density). Similarly, we can represent the signals of all nodes on $\mathcal G$ at time length $T$ as a 3D feature tensor $X \in \mathbb{R}^{T \times N \times D}$. Finally, the traffic prediction problem can be formalized as follows:
\begin{equation}
\label{problem}
[X_{t-P-1}, \cdots, X_{t-1}, X_t; \mathcal G] \xrightarrow{f(\cdot)} [\widehat{X}_{t+1}, \cdots ,\widehat{X}_{t+Q}]
\end{equation}
This formula indicates that given $P$ time lengths of historical observations and graph $\mathcal G$, predict future traffic status for $Q$ time lengths. The task aims to learn a non-linear function $f(\cdot)$ based on the gradient descent of the error. The mathematical form of $\mathcal{L}$ optimization objective is defined as follows:
\begin{equation}
\Theta^* =\arg \min _{\Theta} \ \mathcal{L}(f(A, X_{t-P-1:t}; \Theta), X_{t+1:t+Q})
\end{equation}
where $\Theta$ is the parameters to be optimized in the function.

\section{Graph Learning and Computing}
Although graph neural networks (GNNs) have the advantage of aggregating node neighborhood contexts to generate spatial representations \cite{wu2020comprehensive}, the performance of the task is closely related to the quality of the input graph structure and the computational method used for graph convolution. As shown in Fig. \ref{fig:spatial}, the spatial relationships among nodes in traffic networks are complex and diverse. The complex spatial dependencies behind the traffic system cannot be explored by a single graph design and simple equation \cite{zhu2021survey}. Therefore, this section focuses on the following two issues:
\begin{itemize}
\item \textbf{Q1: How to design a reasonable graph structure? And how to mine underlying spatial relationships from the time-series data itself without prior knowledge?}
\item \textbf{Q2: How to perform efficient convolution computation on the existing graphs?}
\end{itemize}

\subsection{Graph Stucture Learning (Q1)}
Message passing in GNNs is based on local similarity \cite{DBLP:conf/icml/GilmerSRVD17}, where closer nodes exhibit more similar traffic patterns. Most of the spatial-temporal graphs used for traffic prediction use road connection distances or absolute distances of physical coordinates to calculate the weights of edges \cite{DBLP:journals/corr/abs-2104-14917}.  The former is typically a directed graph that reflects the objective distribution of the actual network, while the latter is an undirected graph that only measures the spatial distance between pairs of nodes.

\textbf{Distance-based Graph}. The matrix $A_{D}$ of the distance graph is defined using a threshold Gaussian kernel \cite{DBLP:conf/ijcai/YuYZ18, DBLP:conf/smc/ChenZL22, zhang2022adapgl} as follows:
\begin{equation}
A_{D, ij}=
\begin{cases}
exp(-\frac{d_{i,j}^2}{\sigma^2}),& \text{for}\  i \neq j \  \text{and} \  exp(-\frac{d_{i,j}^2}{\sigma^2}) \ge \epsilon, \\
0, & \text{otherwise}.
\end{cases}
\end{equation}
where $d_{ij}$ is the measured distance of $v_i$ and $v_j$. The threshold $\epsilon$ and the variance $\sigma^2$ are used to control the sparsity and distribution of the matrix $A_D$.

\textbf{Connectivity Graph}, also called as \textbf{Binary Graph} \cite{DBLP:conf/aaai/SongLGW20, DBLP:conf/aaai/GengLWZYYL19, DBLP:journals/tits/LiuCWZLL22}. Similarly, $A_C$ is mathematically defined as follows:
\begin{equation}
A_{C, ij}=
\begin{cases}
1,& \text{if}\ v_i\ \text{connects to}\ v_j, \\
0, & \text{otherwise}.
\end{cases}
\end{equation}

\textbf{Semantic Graph}. We observe that certain nodes are geographically distant but they tend to have the same or similar patterns of traffic variation (They may be lying in the same type of area, such as a residential or commercial region). This suggests that node pairs also have significant semantic correlations. 

Generally, the Dynamic Time Warping (DTW) algorithm \cite{DBLP:conf/cikm/LuGJFZ20, DBLP:conf/kdd/FangLSX21, DBLP:conf/aaai/LiZ21} is used to calculate the similarity in temporal patterns of historical observations. Semantic similarity matrix $A_{SE}$ can be calculated according to the following equation:
\begin{equation}
A_{SE, ij}=
\begin{cases}
1,& \text{DTW}(X^{(i)}, X^{(j)}) \ge \epsilon, \\
0, & \text{otherwise}.
\end{cases}
\end{equation}
where $X^{(i)}$ is the historical observation data of the $i$-th node.

\textbf{Functionality Graph}. The POIs distribution surrounding nodes determines the usage of the district. Studies \cite{DBLP:journals/tits/LvHCCZJ21,DBLP:conf/aaai/GengLWZYYL19, DBLP:conf/ijcai/ShaoJWKXMZZS22} revealed that this composite spatial dependence and heterogeneity largely influence the trend of traffic. In practice, we characterize the region functionality by the category and number of nearby POIs \cite{DBLP:conf/aaai/GengLWZYYL19}, and the formula that defines the functional proximity between node pairs is $A_{F, ij}=\text{sim}(P_{v_i}, P_{v_j}) \in [0, 1]$. Cosine similarity \cite{xia2015learning} is a typical method used to calculate the functional similarity matrix $A_{F}$ with the following equation:
\begin{equation}
A_{F, ij}= \frac{P_{v_i} \cdot P_{v_j}^T}{||P_{v_i}|||P_{v_j}||}
\end{equation}
where $P_{v_i}$ is a vector encoding of POIs for node $v_i$  and its dimensions label the number of categories of POIs. $P_{v_i}[j]$ is calculated in some way to represent the density of POI categories $j$ around node $v_i$. 

\textbf{Distribution Graph}. Some metrics (e.g., Pearson correlation coefficient) can be used to describe the differences in traffic trends between nodes. However, their results are susceptible to the effects of series length. When the series length is small, it is susceptible to noise interference, and when the length is set too large, the variability of the trends is reduced. In contrast, from a macroscopic perspective, it is possible to simultanously combat data noise and effectively measure the overall proximity of nodes by comparing the feature distribution of nodes. KL divergence \cite{DBLP:journals/tit/ErvenH14} and JS divergence \cite{DBLP:conf/aaai/ZhouWXCL20, DBLP:conf/aaai/WangL0W21} are often used to evaluate the similarity of two probability distributions. Let $P_i$ and $P_j$ denote the observed values of two nodes. The distribution matrix $A_{J}$ based on JS divergence (JSD) \cite{DBLP:conf/dasfaa/LiFCZLZ22} can be formulated as follows:
\begin{equation}
JSD(P_i||P_j)= \frac{1}{2}KL(P_i||P_j) + \frac{1}{2}KL(P_j||P_i)
\end{equation}
where $KL(P_i||P_j)$ can be expressed as:
\begin{equation}
KL(P_i||P_j)= \sum_{x \in X}P_{i}(x)log{\frac{P_{i}(x)}{P_{j}(x)}}
\end{equation}

The range of JSD is [0, 1], and smaller values indicate greater distribution similarity. Thus, we define $A_{J, ij}=1-JSD(P_i||P_j)$.

The aforementioned methods of constructing graphs either encode adjacency matrices using prior knowledge or construct similarity matrices based on statistical analysis. They significantly enhance the spatial-temporal awareness ability of the model in auxiliary space modeling, compensating for the information bias introduced by an individual graph.

\begin{table}[h]
\centering
\renewcommand{\arraystretch}{1.5}
\caption{Comparison among different calculation equations of the adaptive matrix.}
\resizebox{0.88\linewidth}{!}{
\begin{tabular}{l|l}
\specialrule{0.12em}{0em}{0em}
\textbf{Method} & \textbf{Equation} \\
\hline
Direct Parametric A & $A=ReLU(W), W\in \mathbb{R}^{N \times N} $ \\
\hline
Undirected Graph A & $A=ReLU(tanh(\alpha(EE^{T}))), E\in \mathbb{R}^{N \times d}$ \\
\hline
Directed Graph A & $A=ReLU(tanh(\alpha(E_{1}E_{2}^{T}))), E_{1}, E_{2}\in \mathbb{R}^{N \times d}$ \\
\hline
Uni-Directed Graph A & $A=ReLU(tanh(\alpha(E_{1}E_{2}^{T} - E_{2}E_{1}^{T}))), E_{1}, E_{2}\in \mathbb{R}^{N \times d}$\\
\hline
Attention-Based A & $A=Softmax(\frac{(<X||E>W_1)(<X||E>W_2)^T}{\sqrt{d}}), E\in \mathbb{R}^{N \times d}$ \\
\specialrule{0.12em}{0em}{0em}
\end{tabular}
}
\label{table:adp}
\end{table}

However, the connectivity relations of pre-defined graphs are often missing and biased. On the one hand, they rely on additional data sources and experience, and on the other hand, it is difficult to depict a spatial dependence panorama. This leads to an inability to extend to general spatial-temporal graph tasks. \textbf{The Adaptive Graph} is based on parameter representations of node embeddings \cite{DBLP:conf/nips/0001YL0020, DBLP:conf/ijcai/WuPLJZ19, tian2021spatial} that are continuously updated during the training phase to reduce model errors. It identifies biases caused by human definitions and captures hidden spatial dependencies.

The adoption of adaptive graphs has made remarkable progress in traffic prediction. Here we organize the frequently used constructive equations in STGNN models \cite{DBLP:conf/kdd/WuPL0CZ20, DBLP:journals/corr/abs-2104-14917}  into Table \ref{table:adp}. These efforts allow the study of graph computation without having to rely on priori knowledge.

\begin{figure}[t]
\centering
\includegraphics[width=\linewidth]{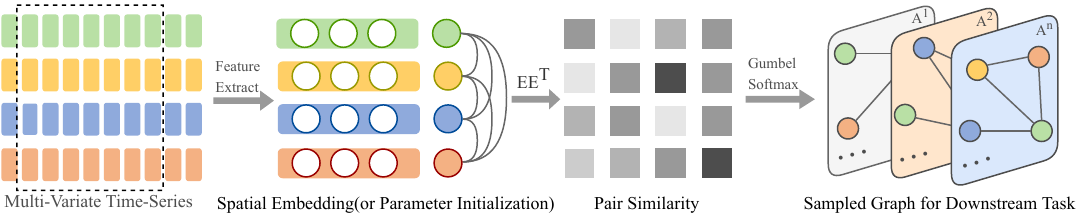}
\caption{The Process of Discrete Graph Structure Learning.}
\label{fig:graoh_learning}
\end{figure}

Meanwhile, a continuous signal sampling-based graph learning and optimization strategy has been proposed \cite{DBLP:conf/iclr/Shang0B21, DBLP:conf/ijcai/YuLYLHWL22}, as illustrated in Fig. \ref{fig:graoh_learning}. It first extracts spatial embedding for each node from the historical observation sequence or initializes embedding parameters directly. Then it computes a pairwise similarity matrix $\Theta$ using the dot product on the spatial embeddings. Finally, it uses the Gumbel softmax trick \cite{DBLP:conf/iclr/JangGP17} to reparameterize the distribution of the probability graph and remove noise information contained in redundant small values. \textbf{The Sampled Graph} is formulated as follows: 
\begin{equation}
A_{ij}= \sigma((log(\theta_{ij}/(1-\theta_{ij}) + (g^1_{ij} - g^2_{ij})/s))),
\end{equation}
where $\Theta$ is a probability matrix, then $\theta_{ij} \in \Theta$ represents the probability of retaining the edge between $v_i$ and $v_j$. Here $g^1_{ij}, g^2_{ij} \sim Gumbel(0, 1)$, $s$ is a temperature hyperparameter.

The sampled graph in the downstream task continuously adapts to the training data to optimize their structure parameters and learn a similarity matrix that minimizes the training error in an end-to-end way \cite{zhu2021survey}. Additionally, a regularization term for the graph error is also added to prevent the learned graph from deviating from the prior graph.

Research on graph learning goes beyond this. Graphs, either learned or pre-defined, can be used as additional information to help models better extract spatial representations. But currently no golden measure of learned graph quality exists, other than prediction accuracy.

\subsection{Graph Computation Method (Q2)}
The essence of GNNs is to aggregate the features of the target node itself and its neighbors to generate high-level hidden representations \cite{wu2020comprehensive}. 
The spectral method \cite{defferrard2016convolutional} uses Chebyshev polynomial approximate filters to achieve and feature extraction of the graph signal. 
The formula is as follows:
\begin{equation}
\Theta \star_{\mathcal{G}} X=\Theta(L) X=\Theta\left(U \Lambda U^{T}\right) X=U \Theta(\Lambda) U^{T} X
\end{equation}
where Graph Fourier Basis $U \in R^{N \times N}$ is the matrix of eigenvectors of the normalized graph Laplacian $L = I_N - D^{-\frac{1}{2}}AD^{-\frac{1}{2}}=U \Lambda U^{T}$.
To balance performance and complexity, in practice, GCN (Graph Convolutional Network) \cite{kipf2016semi} is most commonly used as the $1^{th}$-order approximation of ChebNet. Given the graph signal matrix $X$ and the adjacency matrix $A$, the graph convolutional network can be simplified to the following equation:
\begin{equation}
Z=(I_N+D^{-\frac{1}{2}}AD^{-\frac{1}{2}})XW+b
\end{equation}
where $I_N$ is the identity matrix and $D=diag(\sum^N_{j=0}A_{i,j})$ is the degree matrix. Spatial domain graph convolution is widely used in spatial-temporal graph networks to capture the spatial dependence of undirected graphs.

\textbf{Diffusion Graph Convolution}. On the one hand, the traffic dissemination is directional, while on the other hand, the impact of traffic may come from more distant nodes, making simple GCN inadequate for complex scenarios. As a comparison, diffusion convolution on directed graphs can capture information up to $k$-order bi-directional neighbors, expanding the model's spatial receptive field. The equation \cite{DBLP:conf/iclr/LiYS018, DBLP:conf/ijcai/WuPLJZ19} is expressed as follows:
\begin{equation}
g_{\theta} \star_{\mathcal G} X = \sum^{K-1}_{k=0}(\theta_{k,1}(D_{O}^{-1}A)^k + \theta_{k,2}(D_{I}^{-1}A^T)^k)X,
\end{equation}
where $D_{O}$ and $D_{I}$ are the out-degree and in-degree matrices resp., and $\theta_{k,1}, \theta_{k,2}$ are the learnable parameters.

\textbf{Multi-Hop Graph Convolution}. A basic fact is that cascading GNNs leads to the over-smoothing of signals \cite{DBLP:conf/aaai/LiHW18}, i.e., all nodes converge to the same value. Therefore, the graph convolutional network is typically set to two layers, but shallow networks are unable to capture the rich and deeper spatial features \cite{DBLP:conf/cikm/LuGJFZ20}. The residual connection can mitigate this issue. The computation of a uni-directional multi-hop graph convolution \cite{DBLP:conf/kdd/WuPL0CZ20, DBLP:journals/corr/abs-2104-14917} can be expressed as Eq. (\ref{eq:gcn}) and Eq. (\ref{eq:residual}).
\begin{equation}
\label{eq:gcn}
H^{k+1} =\phi(D^{-1}(A+I)H^{k}W), where \ H^0 = X,
\end{equation}
\begin{equation}
\label{eq:residual}
H^{k+1} = \beta H^0 + (1-\beta)\title{A}H^{k+1},
\end{equation}
where $\phi $ is a signal activation function and $\beta $ is a hyperparameter that controls the proportion of the original state of the root node that is preserved.

The messages from multi-hop nodes are aggregated as the output of the hidden layer using linear weighting or attention aggregation \cite{DBLP:conf/aaai/ChenCXCGF20}, in addition to pooling approaches such as Max and Avg. Formally,
\begin{equation}
H_{out} = \sum^{K-1}_{k=0}\alpha^{(k)}H^{k}.
\end{equation}
The multi-hop graph convolutional network uses multiple layers of convolutions to effectively extract features of the layered local substructures of nodes.

\textbf{Graph Attention Network.} The graph attention \cite{velivckovic2017graph}, which dynamically calculates edge weights between nodes based on feature similarity, is more suitable for real-time changing traffic scenarios. And compared to GCN, GAT is more flexible. Let the feature vectors of $v_i$ and $v_j$ be $h_i$ and $h_j \in \mathbb{R}^{D}$ , and $\mathcal N_i$ be the set of neighbors to $v_i$. The equation of the graph attention is as follows:
\begin{equation}
e_{ij}=a(Wh_i, Wh_j), j \in \mathcal N_i,
\end{equation}
\begin{equation}
\alpha_{ij}=softmax(e_{ij})=\frac{exp(LeakyReLU(e_{ij}))}{\sum_{k \in \mathcal N_i}exp(LeakyReLU(e_{ik}))},
\end{equation}
where $W \in \mathbb{R}^{D \times F}$ is a learnable linear matrix, and $a: \mathbb{R}^F \times \mathbb{R}^F \rightarrow \mathbb{R}$ maps the combined parameter matrix into a scalar. $a_{ij}$ denotes the normalized attention score.

To obtain more abundant representations, the $K$-head attention performs multiple transformations of independent subspaces before concatenating them to obtain the calculated result:
\begin{equation}
h_i'={||}^K_{k=0}\sigma(\sum_{j \in N_i}\alpha_{ij}W^kh_j).
\end{equation}
Formally, we formulate the above process in a unified equation:
\begin{equation}
H^{l}=(\tilde{A} \odot M)H^{l-1}W,
\end{equation}
where $\tilde{A} = A + I_N$, $\odot$ is the element-wise product, $M \in \mathbb{R}^{N \times N}$ denotes the dynamic attention matrix. $H^l \in \mathbb{R}^{N \times F}$ is the $l$-th head graph attention layer output, and when $l = 0$, $H^0=X$. 

The exploration of deeper spatial features continues to be an ongoing area of research. Currently, the above graph computation methods have been widely adopted in modeling spatial dependencies for STGNNs.

\section{Spatial-Temporal Graph Neural Networks}

\begin{figure}[t]
\centering
\includegraphics[width=\linewidth]{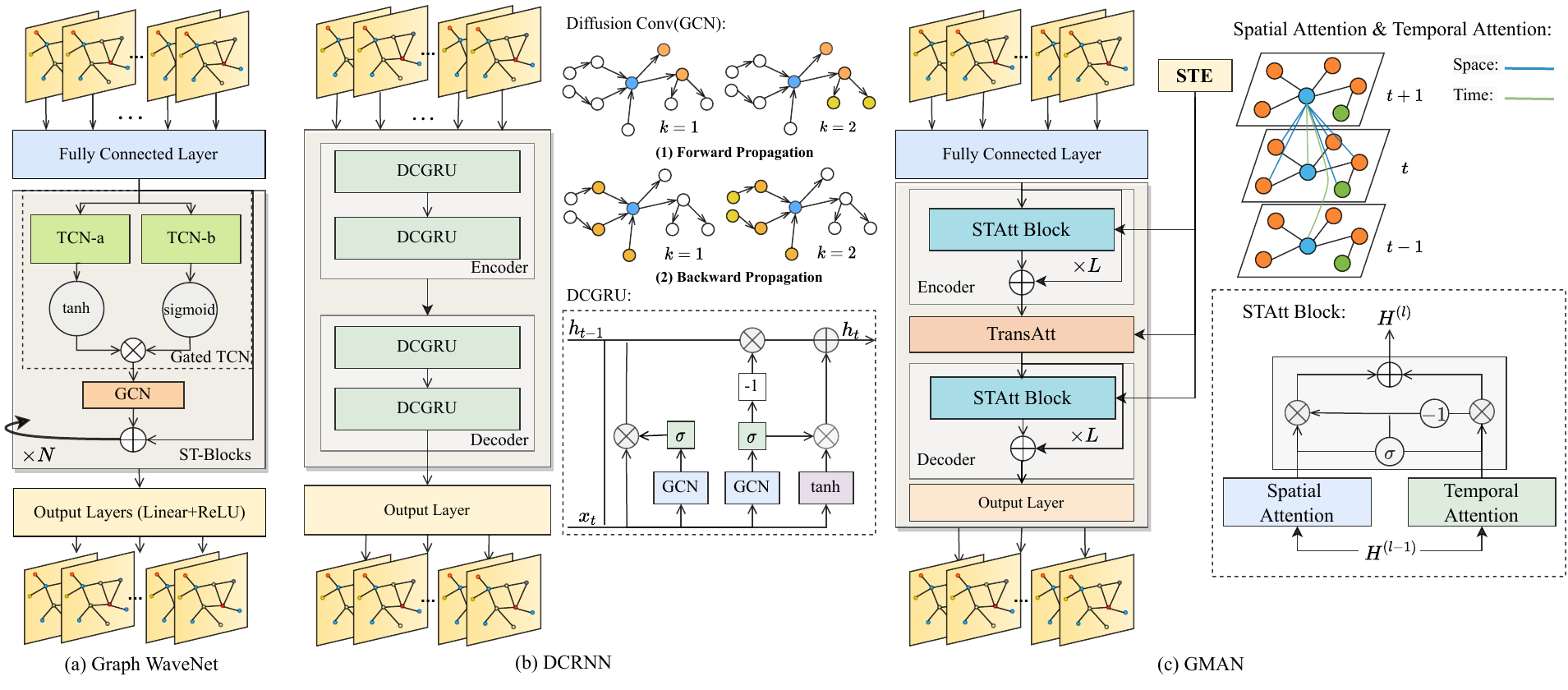}
\caption{Architectures of Representative STGNN Models.}
\label{fig:STGNN_models}
\vspace{-4mm}
\end{figure}

Spatial-temporal graph neural networks (STGNNs) have gained popularity as a deep learning approach that integrates graph convolutional layers into sequence models. This methodology effectively captures the spatial and temporal characteristics of traffic signals. By considering the entire road network and modeling spatial information, STGNNs surpass the limitations of analyzing independent data streams separately and the prediction accuracy of the model is significantly improved.
Through the rapid development in the past five years, a large amount of work has been accumulated to apply spatial-temporal graphs to traffic prediction. According to the modeling strategy of the temporal axis, it can be divided into three categories, namely CNN-Based, RNN-Based, and Attention-Based. The representative STGNNs architectures are shown in Fig. \ref{fig:STGNN_models}. The CNN-Based methods (STGCN, Graph WaveNet, MTGNN, etc.) employ 1D CNNs (TCNs) in tandem with graph convolutional layers to construct ST-Blocks and then learn asynchronous spatial-temporal patterns through the cascading ST-Blocks. 1D CNNs capture more long-range temporal features by stacking convolutional layers or adding dilation factors, thus enjoying the advantages of good computational efficiency and gradient stabilization. However, the implicit temporal connections represented by fixed convolution kernels deprive them of some flexibility and, more importantly, fail to capture the synchronized information in spatial-temporal signals. 

In contrast, Recurrent Neural Networks (RNNs, e.g., LSTM, GRU) are powerful in modeling sequence dependencies. Many RNN-Based methods (DCRNN, MRA-BGCN, AGCRN, etc.) extend the fully-connected operation in RNNs using GCNs so that they use graph convolution to capture local spatial dependencies at both input-to-state and state transitions. This design approach associates each time step with graph convolution, enabling the learning of spatial-temporal signals that undergo synchronous changes. However, it suffers from gradient instability and being very time-consuming in training and inference stage. Furthermore, the forgetten gate mechanism in RNNs has constrained their capability in capturing long-term temporal dependencies.

Theoretically, the attention mechanism and its variants possess a large global receptive field, which has led to their widespread use in capturing long-term temporal dependencies in sequence models. Attention-based methods (such as GMAN, ASTGCN, DSTAGNN, etc.) typically combine temporal attention with spatial attention, excelling at handling global contextual features and spatial-temporal correlations in the evolution of states. By dynamically computing attention scores, these methods enhance the model's focus on critical information, effectively addressing the limitations of TCN and RNN approaches. However, the attention mechanism introduces additional parameters and complexity, making the model overly sensitive to noise or irrelevant information in the training data, which can impact its generalization capability. In addition to the aforementioned methods, there are various other widely-used time series processing techniques, including neural controlled differential equations, Fourier domain transformations, and time pattern decomposition, among others. These provide a diverse set of tools and techniques that can be employed for analyzing and modeling different types and characteristics of time series data.

In recent studies, ``GRU-GCN" is one of the most used spatial-temporal graph modeling frameworks. To overcome the challenges of high run-time and error accumulation, several works have proposed ``curriculum learning" to optimize the training phase of the model. 
Curriculum Learning \cite{DBLP:conf/kdd/WuPL0CZ20, DBLP:journals/corr/abs-2104-14917} argues that it is not necessary to calculate the error and backpropagation for all time steps early in the training but to gradually increase the prediction length of the model as the number of iterations increases, i.e., in a progressive convergence way. 
This strategy for the encoder-decoder architecture substantially reduces the training time consumption and alleviates the pressure in terms of efficiency and resource occupation. 

\begin{table}[t]
\renewcommand{\arraystretch}{1.70}
\caption{Summary of Representative Models for Spatial and Temporal Modeling Techniques.}
\centering
{\fontsize{8}{6.4}\selectfont
\resizebox{1.00\linewidth}{!}{
\begin{tabular}{|c|c|c|c|c|}
\specialrule{0.1em}{0em}{0em}
\textbf{Venue} & \textbf{Model} & \textbf{Graph Construction} & \textbf{Spatial Components} & \textbf{Temporal Components}\\
\specialrule{0.02em}{0em}{0em}
IJCAI 18 & STGCN \cite{DBLP:conf/ijcai/YuYZ18} & Distance-based Graph & ChebNet/GCN & Gated TCN\\
ICLR 18 & DCRNN \cite{DBLP:conf/iclr/LiYS018} & Distance-based Graph & DGC & GRU\\
AUAI 18 & GaAN \cite{DBLP:conf/uai/ZhangSXMKY18} & Distance-based Graph & GAT & GRU\\
IJCAI 19 & GWNET \cite{DBLP:conf/ijcai/WuPLJZ19} & Distance-based/Adaptive Graph & DGC & Dilated TCN\\
AAAI 19 & ASTGCN \cite{DBLP:conf/aaai/GuoLFSW19} & Connectivity Graph & GCN; Attention & TCN; Attention\\
AAAI 19 & ST-MGCN \cite{DBLP:conf/aaai/GengLWZYYL19} & Multiple Graph & ChebNet & RNN\\
KDD 19 & ST-MetaNet \cite{DBLP:conf/kdd/PanLW00Z19} & Distance-based graph & Meta-GAT & Meta-GRU\\
IJCAI 20 & LSGCN \cite{DBLP:conf/ijcai/HuangHLDK20} & Distance-based Graph & ChebNet; GAT & Gated TCN\\
AAAI 20 & STSGCN \cite{DBLP:conf/aaai/SongLGW20} & Connectivity Graph & GCN & GCN\\
AAAI 20 & GMAN \cite{DBLP:conf/aaai/ZhengFW020} & Distance-based Graph & Embedding; Attention & Embedding; Attention\\
KDD 20 & MTGNN \cite{DBLP:conf/kdd/WuPL0CZ20} & Adaptive Graph & MHGC & Dilated TCN \\
NIPS 20 & AGCRN \cite{DBLP:conf/nips/0001YL0020} & Adaptive Graph & GCN & GRU\\
CIKM 20 & STAG-GCN \cite{DBLP:conf/cikm/LuGJFZ20} & Multiple Graph & GCN; GAT & TCN; Self-Attention \\
TITS 20 & T-MGCN \cite{DBLP:journals/tits/LvHCCZJ21} & Multiple Graph & GCN & GRU \\
AAAI 20 & MRA-BGCN \cite{DBLP:conf/aaai/ChenCXCGF20} & Distance-based/Edge Graph & MHGC & GRU \\
WWW 20 & STGNN \cite{DBLP:conf/www/Wang0WJWTJY20} & Distance-based Graph & GAT & GRU; Transformer\\
ICLR 21 & GTS \cite{DBLP:conf/iclr/Shang0B21} & Sampled Graph & DGC & GRU\\
TKDD 21 & DGCRN \cite{DBLP:journals/corr/abs-2104-14917} & Dynamic Graph & MHGC & GRU\\
TKDE 21 & ASTGNN \cite{DBLP:journals/tkde/GuoLWLC22} & Distance-based Graph & GAT & TCN; Transformer\\
AAAI 21 & STFGNN \cite{DBLP:conf/aaai/LiZ21} & Semantic Graph & GCN & GCN; Dilated TCN\\
AAAI 21 & CCRNN \cite{DBLP:conf/aaai/YeSDF021} & Adaptive Graph & GCN & GRU\\
KDD 21 & STGODE \cite{DBLP:conf/kdd/FangLSX21} & Distance-Based/Semantic Graph & GCN; ODE & TCN \\
KDD 21 & DMSTGCN \cite{DBLP:conf/kdd/HanDSFL021} & Dynamic Graph & DGC & Dilated TCN\\
IJCAI 22 & RGSL \cite{DBLP:conf/ijcai/YuLYLHWL22} & Connectivity/Sampled Graph & GCN & GRU\\
KDD 22 & ESG \cite{DBLP:conf/kdd/YeL0SLF022}& Dynamic Graph & GRU; DGC & Dilated TCN\\
KDD 22 & STEP \cite{DBLP:conf/kdd/ShaoZWX22}& Sampled Graph & DGC & Transformer; Dilated TCN\\
AAAI 22 & STG-NCDE \cite{DBLP:conf/aaai/0002CHP22} & Adaptive Graph & GCN; NCDE & NCDE\\
ICML 22 & DSTAGNN \cite{DBLP:conf/icml/LanMHWYL22} & Dynamic Graph & ChebNet; Attention & Gated TCN; Attention\\
\specialrule{0.1em}{0em}{0em}
\end{tabular}
}
}
\begin{flushleft}
\small
\footnotesize
\textbf{*Note}: To simplify the presentation, we use the terms \textbf{``DGC"} to refer to Diffusion Graph Convolution and \textbf{``MHGC"} to refer to Multi-Hop Graph Convolution. We refer to a model with three or more graphs as a \textbf{``Multiple Graph"} model.
\end{flushleft}
\label{tab:stgnn}
\vspace{-4mm}
\end{table}
 
Just like the No-Free-Lunch theorem \cite{wolpert2002supervised} in machine learning, using STGNNs to model the spatial-temporal correlation of traffic scenes requires selecting appropriate components according to specific problems and conditions. Here we summarize some representative research works, as shown in Table \ref{tab:stgnn}. Some of these methods had reached the state-of-the-art in prediction tasks. STGCN \cite{DBLP:conf/ijcai/YuYZ18} first combines gated temporal convolution with ChebNet operator for spatial-temporal prediction and achieves better performance than traditional time series models in all metrics. DCRNN \cite{DBLP:conf/iclr/LiYS018} extends the diffusion graph convolution to a recurrent neural network of encoder-decoder to solve the directional problem of asymmetric traffic graph propagation. Graph WaveNet \cite{DBLP:conf/ijcai/WuPLJZ19} innovatively proposes an adaptive graph dissipation of biases caused by human-defined spatial relations, and WaveNet based on causal convolution is used to learn temporal relations. Inspired by ST-ResNet \cite{DBLP:conf/aaai/ZhangZQ17}, ASTGCN \cite{DBLP:conf/aaai/GuoLFSW19} proposes a set of temporal components called Clonesss, Period and Trend. Structurally, it utilizes a skip connection to connect the spatial-temporal attention layer to the convolutional layer to form model branchs, and finally fuses the three components together for prediction. These early studies of spatio-temporal graph neural networks significantly outperform traditional statistical methods and machine learning models in terms of predictive performance.

The complexity of traffic scenarios has also given rise to some novel research perspectives. STSGCN \cite{DBLP:conf/aaai/SongLGW20} argues that spatial-temporal dependence often affects traffic volumes not individually but synergistically. It proposes a unique local spatial-temporal graph for capturing spatial-temporal heterogeneity and models synchronous spatial-temporal relationships through multiple graph convolutional layers. GMAN \cite{DBLP:conf/aaai/ZhengFW020} proposes gated spatial-temporal attention fusion to capture dynamic nonlinear spatial-temporal correlations and establish the temporal connections between historical and future time steps based on transforming attention. STFGNN \cite{DBLP:conf/aaai/LiZ21} combines a novel fusion operation to learn hidden dependencies from spatial and temporal graphs, and handles long sequences by stacking fusion graphs and gated convolutional modules. However, the accuracy of these methods heavily relies on pre-defined graph designs, and the number of layers of graph convolution is superficial.

Adaptive graph convolutional recurrent network (AGCRN) \cite{DBLP:conf/nips/0001YL0020}  designs two types of adaptive modules based on parameter decomposition. Firstly, the node adaptation module decomposes the shared weights and biases to generate node-specific parameters to capture node-specific patterns. Secondly, the data-adaptive graph generation module automatically infers the interdependencies between different traffic series. STGNN \cite{DBLP:conf/www/Wang0WJWTJY20} first combines the improved GAT and GRU, and then captures different scales of temporal patterns by concatenating a Transformer neural network architecture. This approach can effectively learn the dependencies among spatio-temporal data and provides a design framework that can be readily adapted. ASTGNN \cite{DBLP:journals/tkde/GuoLWLC22} adopts a Transformer-like spatio-temporal encoding-decoding architecture. It first extends the computation of query, key, and value matrices using temporal convolutions, proposes a trend-aware attention layer, and then replaces the feedforward network layer with GAT. The method has been proven effective on multiple traffic datasets, but the complex parameter training brings heavy system strain. DGCRN \cite{DBLP:journals/corr/abs-2104-14917} found the fact that the connectivity of nodes is not immutable, but dynamically evolves with time periods. It proposes using a hyper-network to generate a dynamic adjacency matrix before each step of the RNN to accommodate the dynamic changes in the road network. Additionally, the generated dynamic matrix is merged with the original road network matrix to capture more spatial information. DMSTGCN \cite{DBLP:conf/kdd/HanDSFL021} designs a dynamic graph constructor and dynamic graph convolution method to propagate node hidden states based on dynamic spatial relationships. It also provides a multi-aspect fusion module to merge auxiliary hidden states and primary hidden states in both time and space. They have made a lot of efforts in graph design and graph computing. 

Spatio-Temporal Multi-Graph Convolution Network (ST-MGCN) \cite{DBLP:conf/aaai/GengLWZYYL19}  and Temporal Multi-Graph Convolutional Network (T-MGCN) \cite{DBLP:journals/tits/LvHCCZJ21} devise multiple attribute graphs to assist in enhanced spatial modeling and mine spatial information from multiple insights. But this means that more expert knowledge is required. STGODE \cite{DBLP:conf/kdd/FangLSX21} adopts ODE to handle multipe layer GCN over-smoothing problem by expressing residual-connected GCNs as continuous GCNs. Then it adopts the dual branching of TCN and CGCN to solve the spatial-temporal prediction problem. Similarly, STG-NCDE \cite{DBLP:conf/aaai/0002CHP22} employs neural control differential equations to tackle the knowledge of temporal and spatial dimensions separately. The neural differential equation restores the temporal continuity that is lost due to interval sampling. Unlike previous work, STEP \cite{DBLP:conf/kdd/ShaoZWX22} is motivated by unsupervised learning and proposes a time-series pre-training strategy to enhance the graph structural design of STGNNs with promising results.

According to our survey, the studies of spatial-temporal graphs focus on designing graph structures and combining spatial-temporal components. Meanwhile, other related pieces of techniques (unsupervised pre-training, generative adversarial networks, graph contrastive learning, reinforcement learning, etc.) are widely applied to traffic prediction, which greatly expand the technical basis of spatial-temporal data mining and achieve great performance gains.

\section{Benchmark And Evaluation}
In this section, we first provides an overview of the datasets and models, the experimental setup, and the evaluation results used in the benchmark. Then we analyze the performance and efficiency of some models through the visualization of charts. Finally, we provide a brief introduction to the benchmark interface and its extended usage.

\begin{table}[!h]
\renewcommand{\arraystretch}{1.4}
\caption{The overall information about the datasets used in the benchmark.}
\centering
\resizebox{0.70\linewidth}{!}{
\begin{tabular}{cccccc}
\toprule[1.2pt]
\multicolumn{1}{l}{\textbf{Traffic Type}} & \textbf{Datasets} & \textbf{Nodes} & \textbf{Edges} & \textbf{Time Steps} & \textbf{Missing Ratio} \\ \hline
\multirow{2}{*}{Speed}           & METR-LA & 207 & 1515 & 34,272 & 8.109\% \\ 
                                 & PEMS-BAY & 325 & 2369 &
                                 52,116  & 0.003\% \\ \hline
\multirow{2}{*}{Flow}            & PEMSD4 & 307 & 340 & 16,992 & 3.182\% \\ 
                                 & PEMSD8 & 170 & 295 & 17,856 & 0.696\% \\
\bottomrule[1.2pt]
\end{tabular}
}
 \label{tab:datasetlabel}
\end{table}

\subsection{Benchmark Implementation}
In view of the heterogeneity of the traffic data, we selected two kinds of datasets, traffic speed (METR-LA, PEMS-BAY) \cite{DBLP:conf/ijcai/WuPLJZ19, DBLP:conf/iclr/LiYS018} and traffic flow (PEMSD4, PEMSD8) \cite{DBLP:conf/aaai/SongLGW20, DBLP:conf/nips/0001YL0020}, for the building of the benchmark. They are both sampled at 5-minute intervals, and the detailed statistical information of the data is shown in Table \ref{tab:datasetlabel}. At the same time, we selected some representative spatial-temporal graph models in these datasets for comparative studies. Our experiments are conducted on a GPU server with eight GeForce GTX 1080Ti graphics cards, using the unified deep learning framework PyTorch 1.8.0. The raw data are standardized using Z-Score \cite{cheadle2003analysis}. To maintain consistency with previous studies, we divide the speed data into the training set, validation set, and test set in the ratio of 7:1:2, and the flow data in the ratio of 6:2:2. If the validation error converges within 15-20 epochs or stops after 100 epochs, the training model would stop early and the best model on the validation data is saved \cite{DBLP:journals/corr/abs-2302-12598}. In the multi-step prediction task, we set both $P$ and $Q$ of the problem definition (\ref{problem}) to 12. For the specific model parameters and settings including optimizer, learning rate, loss function and model parameters, we are faithful to the original paper on the one hand and conduct several parameter tuning efforts to obtain better experimental results on the other hand. 

In our experiments, we evaluate the model results using the Mask-Based Root Mean Square Error (RMSE), Mean Absolute Error (MAE), and Mean Absolute Percentage Error (MAPE) as metrics, where the zero values (i.e., noisy data) will be ignored \cite{DBLP:conf/cikm/JiangYWWDLCDSS21}. Their initially defined equations are as follows:
\begin{equation}
\resizebox{0.30\linewidth}{!}{$  \text{MAE}=\frac{1}{M}\sum^M_{i=1}|Y_i - \widehat{Y_i}|, \ \ $}
\resizebox{0.30\linewidth}{!}{$  \text{RMSE}=\sqrt{\frac{1}{M}\sum^M_{i=1}(Y_i - \widehat{Y_i})^2}, \ \ $}
\resizebox{0.30\linewidth}{!}{$  \text{MAPE}=\frac{100\%}{M}\sum^M_{i=1}\left|\frac{Y_i - \widehat{Y_i}}{Y_i}\right|.$}
\end{equation}
where $M$ is the number of values to predict, $Y_i$ is the prediction result and $\widehat{Y_i}$ is the ground truth. The smaller their values, the better the performance of the method is indicated.

\begin{table}[t]
 \renewcommand{\arraystretch}{0.78}
 \centering
 \caption{Performance of Spatial-Temporal Graph Neural Networks for Multi-Step Prediction on Traffic Speed.}
 { 
 \resizebox{140mm}{32mm}{
 \begin{tabular}{c c c c c c c c c c c c c }
  \specialrule{0.12em}{0em}{0.2em}
  \multirow{2}{*}{Datasets} & \multirow{2}{*}{Models} & \multicolumn{3}{c}{Horizon 3} & &\multicolumn{3}{c}{Horizon 6} & & \multicolumn{3}{c}{Horizon 12} \\
  \cline{3-5} \cline{7-9} \cline{11-13} 
  {} & {}& \multicolumn{1}{c}{MAE} & \multicolumn{1}{c}{RMSE} & \multicolumn{1}{c}{MAPE} & & \multicolumn{1}{c}{\makecell[c]{MAE \\[0.pt]}} & \multicolumn{1}{c}{RMSE} & \multicolumn{1}{c}{MAPE} & & \multicolumn{1}{c}{MAE} & \multicolumn{1}{c}{RMSE} & \multicolumn{1}{c}{MAPE}\\
  \specialrule{0.05em}{0em}{0.5em}
  \multirow{8}{*}{METR-LA} & T-GCN & 3.09 & 5.77 & 8.81\% & & 3.60 & 6.90 & 10.61\% & & 4.29 & 8.39 & 12.56\% \\ 
  {} & GMAN & 2.84 & 5.71 & 7.54\% & & 3.16    & 6.65  & 8.79\%  & & 3.50 & 7.46 & 10.18\% \\
  {} & STGCN & 2.79 & 5.34  & 7.29\% & & 3.21  & 6.48  & 8.81\%  & & 3.74 & 7.72 & 10.71\% \\ 
  {} & DCRNN & 2.77 & 5.37  & 7.16\% & & 3.14  & 6.43  & 8.56\%  & & 3.59 & 7.57 & 10.32\% \\ 
  {} & GTS & 2.75 & 5.26  & 7.12\% & & 3.13  & 6.29  & 8.59\%  & & 3.56 & 7.33 & 10.21\% \\ 
  {} & GWNET & 2.69 & 5.14  & 6.94\% & & 3.06  & 6.15  & 8.25\%  & & 3.52 & 7.30 & \textbf{9.80\%} \\ 
  {} & MTGNN & 2.68 & 5.16  & 6.91\% & & 3.03  & 6.14  & 8.29\%  & & \textbf{3.46} & 7.20 & 9.87\% \\ 
  {} & DGCRN & \textbf{2.63} & \textbf{5.01}  & \textbf{6.78\%} & & \textbf{2.99}  & \textbf{6.01}  & \textbf{8.09\%}  & & 3.48 & \textbf{7.18} & 9.90\% \\ 
    \specialrule{0.05em}{0.1em}{0.4em}
  \multirow{8}{*}{PEMS-BAY} & T-GCN & 1.49 & 2.98 & 3.13\% & & 1.86 & 3.99 & 4.18\% & & 2.29 & 5.01 & 5.46\% \\ 
  {} & STGCN & 1.39 & 2.92  & 2.99\% & & 1.75  & 3.95  & 4.00\%  & & 2.11 & 4.81 & 5.06\% \\ 
  {} & GMAN & 1.36 & 2.90 & 2.99\% & & 1.64    & 3.79  & 3.80\%  & & \textbf{1.88} & \textbf{4.32} & 4.50\% \\
  {} & DCRNN & 1.35 & 2.84  & 2.84\% & & 1.69  & 3.87  & 3.80\%  & & 2.01 & 4.69 & 4.69\% \\ 
  {} & GTS & 1.33 & 2.80  & 2.82\% & & 1.64  & 3.77  & 3.77\%  & & 1.93 & 4.49 & 4.59\% \\ 
  {} & GWNET & 1.31 & 2.75  & 2.72\% & & 1.65  & 3.70  & 3.74\%  & & 1.98 & 4.55 & 4.64\% \\ 
  {} & MTGNN & 1.33 & 2.80  & 2.73\% & & 1.65  & 3.73  & 3.65\%  & & 1.94 & 4.49 & 4.56\% \\ 
  {} & DGCRN & \textbf{1.29} & \textbf{2.70}  & \textbf{2.66\%} & & \textbf{1.62}  & \textbf{3.65}  & \textbf{3.53\%}  & & 1.92 & 4.42 & \textbf{4.40\%} \\ 
  \specialrule{0.12em}{0em}{0em}
 \end{tabular}
 }
 }
 \label{tab:result1}
\end{table}

\begin{table}[!t]
\renewcommand{\arraystretch}{1.32}
 \centering
 \caption{Average Performance of Spatial-Temporal Graph Neural Networks on Traffic Flow.}
 \resizebox{140mm}{16mm}{
\begin{tabular}{c|c|c c c c c c c c c c}
\specialrule{0.12em}{0em}{0em}
Datasets                & Metrics & MSTGCN & ASTGCN & STSGCN & STGODE & GMSDR & AGCRN & RGSL & MTGNN & STG-NCDE & DAAGCN \\ \hline
\multirow{3}{*}{PEMSD4} & MAE     &  23.58 &  21.83 &  21.16 &  21.04 & 19.79 &  19.67 &  19.29 &  19.22 &    19.21 &  \textbf{18.78} \\ \cline{2-2}   & RMSE    & 36.86  &  34.36  &  34.13   &  33.46   &  32.04  &  32.31  & 31.48  &  31.28 &  31.09  & \textbf{30.79}  \\ \cline{2-2}  & MAPE  &  15.85\%  &  14.13\%  &  14.39\%  &  14.14\%  &     13.34\%  &  12.91\%  &  12.65\% &     12.64\%  &  12.79\%   & \textbf{12.23\%}       \\ \hline
\multirow{3}{*}{PEMSD8} & MAE  &
18.25 &  18.10  &  17.13  & 16.81  & 16.07 &  15.81  &  15.60  & 15.56  &  15.45 &  \textbf{15.06}  \\ \cline{2-2} 
& RMSE  &  28.29  &  28.06 &  26.47 & 26.44  &  25.15   &  25.04 & 25.07  &  24.90   &  24.81 &  \textbf{24.36}  \\ \cline{2-2}  & MAPE  &  11.57\%  &  11.13\%  &  11.15\%  &  10.57\%  & 10.41\%  &  10.19\%  &  10.04\%  &    9.83\% &  9.92\%   & \textbf{9.74\%}       \\ \specialrule{0.12em}{0em}{0em}
\end{tabular}
 \label{tab:result2}
}
\end{table}

Table \ref{tab:result1} reports the multi-step prediction results of STGNNs for 15 minutes, 30 minutes, and 1 hour in the traffic speed datasets. Table \ref{tab:result2} records the average performance of the model across all time steps in the traffic flow datasets. It can be noticed that all our reproduced results are very close to the results reported in the original paper. We can draw the following conclusions from the abundant data information: In both METR-LA and PEMS-BAY, DGCRN achieves state-of-the-art performance in almost all horizons except a few metrics. The overall prediction accuracy of GWNET and MTGNN are similar. Both of them use an adaptive graph learning strategy, which is superior to some graph methods based on distance representation. We can also observe that GMAN performs worse at short-term horizons and better at longer times (Horizon 12), suggesting that spatial-temporal attention helps to improve long-range predictions.

DAAGCN is a dynamic spatial-temporal recurrent network combined with GAN, which achieves the best performance on both PEMSD4 and PEMSD8. MTGNN still performs well on traffic flow data, indicating its robustness and versatility across both speed and flow datasets. In our experiments, we find that AGCRN not only achieves higher prediction accuracy compared to ASTGCN, STSGCN, and STGODE, but also occupies very few system resources in training stage. RGSL is an improvement on AGCRN, which proposed to fuse explicit and implicit graphs to effectively model node pair dependencies. STG-NCDE employs nonlinear differential equations to describe the continuous dynamic evolution of node features in both time and space. Despite its superior performance compared to the majority of methods, the complex differential operators increase both the training and testing time of the model. Integrating the experimental data from the two tables, we can find that over the past five years, the performance of the STGNN (Spatial-Temporal Graph Neural Network) models on traffic prediction tasks has improved by 15\% to 20\%. This indicates that the models have made significant progress in capturing more in-depth spatial-temporal patterns. In terms of method design, STGNNs have gradually moved away from relying on prior graphs, and have instead attempted to establish a more universal and effective spatial-temporal prediction model.

\begin{figure}[!t]
\centering
\subfloat{\includegraphics[width=4.2in]{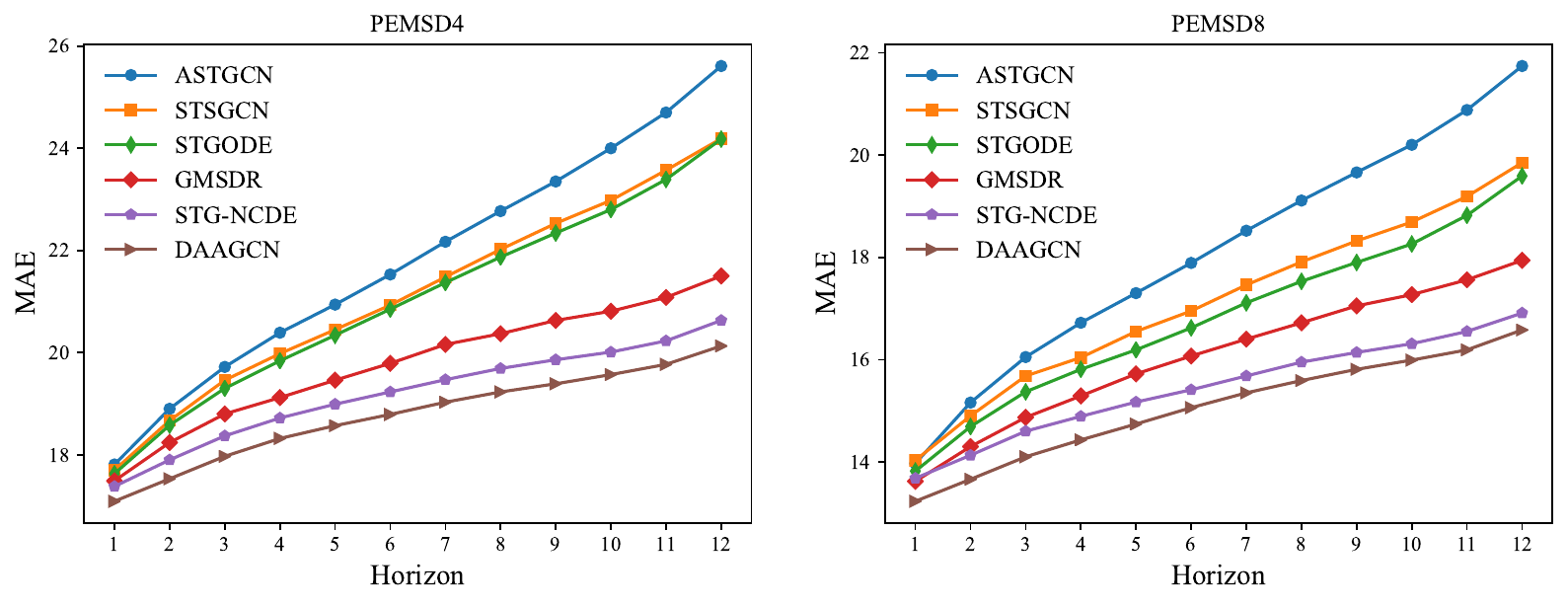}%
\label{a1}}
\hfill
\subfloat{\includegraphics[width=4.2in]{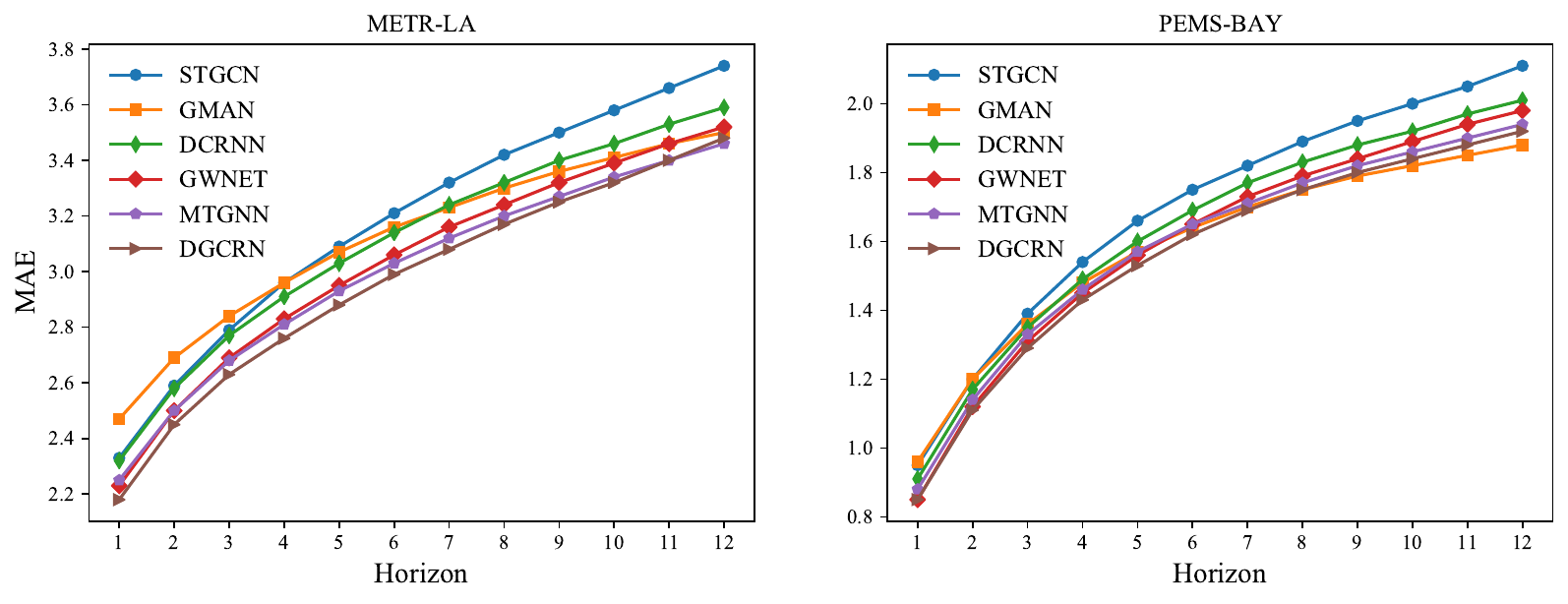}%
\label{a2}}
\caption{Visualization of the MAE for each prediction step.}
\label{fig:summary2}
\end{figure}

To provide a more intuitive comparison of the performance differences of different methods across multiple prediction horizons, we plotted the MAE line charts for Horizons 1 to 12 in Fig. \ref{fig:summary2}. Through this figure, we can clearly observe the performance differences of these models in short-term and long-term forecasting. As the time interval increases, the model's uncertainty about the future traffic network increases. Each step of multi-step prediction depends on the previous prediction, and the error will gradually accumulate and amplify, so the prediction error of all models will gradually increase. On the PEMS traffic flow data, we found that ASTGCN (blue line), STSGCN (orange line), and GMSDR (red line) have significant differences in the prediction range of Horizon 6 to 12, which reflects their different modeling capabilities for long-term prediction. GMSDR effectively mitigates the error propagation in multi-step prediction. On METR-LA and PEMS-BAY, we observed relatively smaller changes in the multi-step prediction error rates of the models. Therefore, the key to overall MAE reduction is to improve the model's ability to handle long-term spatio-temporal dependencies.

\begin{figure}[!t]
\centering
\subfloat{\includegraphics[width=2.6in]{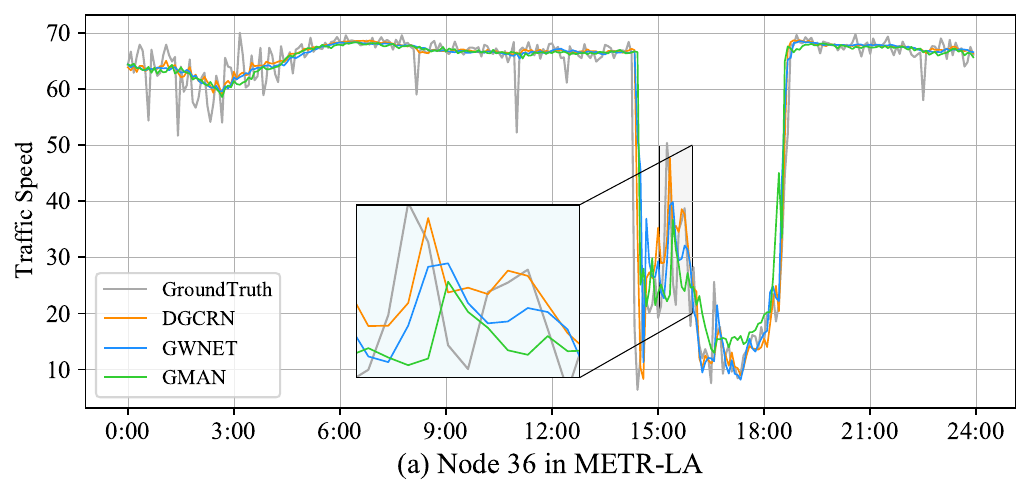}%
\label{a}}
\hfill
\subfloat{\includegraphics[width=2.6in]{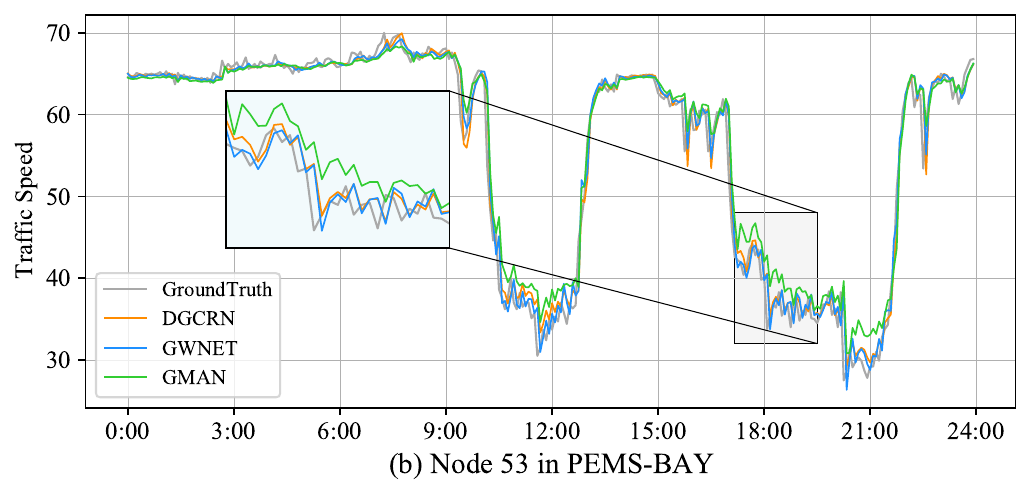}%
\label{b}}
\vspace{-2mm}
\subfloat{\includegraphics[width=2.6in]{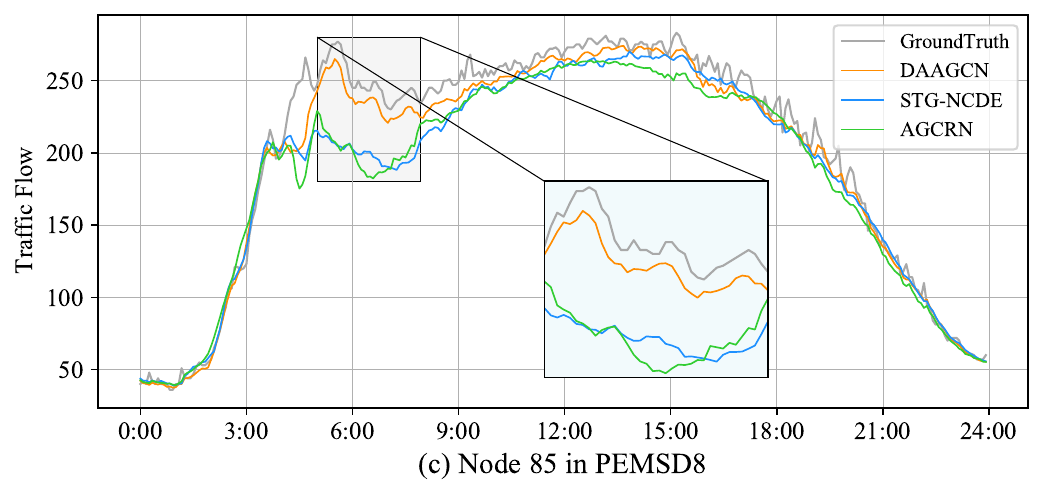}%
\label{c}}
\hfill
\subfloat{\includegraphics[width=2.6in]{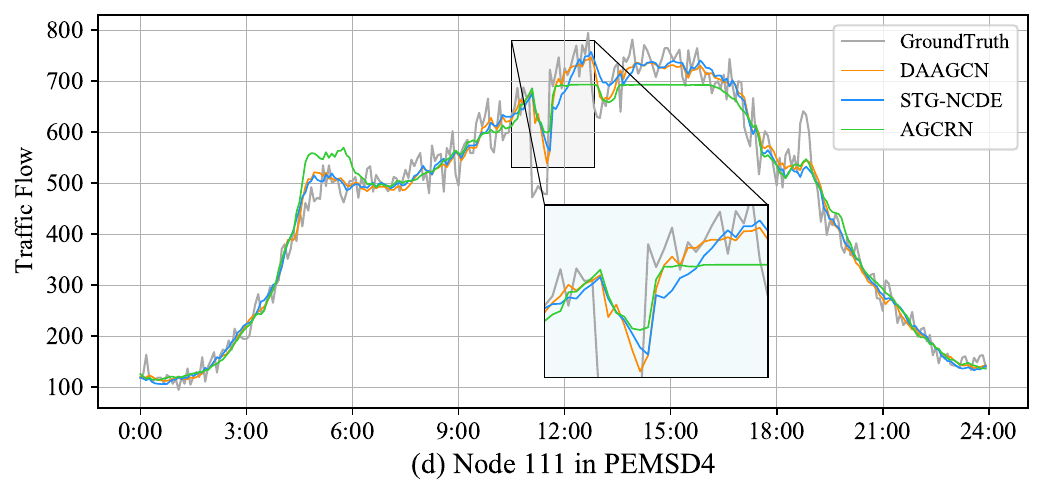}%
\label{d}}
\caption{Traffic speed or flow prediction at timeslots in one day on different datasets.}
\label{fig:visualization}
\end{figure}

\subsection{Case Study}
We plot the time-series curve of ground truth versus model prediction for a random selection of road network sensors from the four datasets for a given day as in Fig. \ref{fig:visualization}. For better presentation, a limited number of models are selected for comparison across the datasets, rather than all models listed in Table \ref{tab:result1}. We observe that: (1) Node 36 in METR-LA has a sudden decrease in the speed of vehicle traffic during the 15:00-18:00 segment, which indicates that the highway is jammed. Node 53 in PEMS-BAY shows successive traffic congestion during the hours of 10:00-12:00 and 18:00-21:00. (2) All GMAN, GWNET, and DGCRN have the capability of capturing stable temporal patterns in the data and can fit the traffic trends in non-congested periods better. This demonstrates the excellent performance of STGNNs in capturing temporal and spatial dependencies. (3) Under some traffic congestion conditions, they learn the valley and peak trends. The fitting effect of DGCRN is more prominent compared to GMAN and GWNET, because it seems to be more sensitive to complex traffic state changes. The predicted values of future traffic data by STGNNs show a lag effect compared to the actual values. This issue needs to be addressed in future research on modeling temporal correlations.


Similarly, we selected DAAGCN, AGCRN and STG-NCDE for comparison in PEMSD4 and PEMSD8. The state of the flow data to change over time is not as smooth as the speed, and there are many bumps on the surface of the curve, which makes traffic flow prediction more challenging. Node 85 in PEMSD8 is the peak of traffic during the 5-8h time frame. DAAGCN is capable of learning the shifting spatial-temporal patterns, while AGCRN and STG-NCDE clearly deviate from the ground truth of traffic. In addition, in PEMSD4, Node 111 shows a sharp decrease in traffic between 11-12h, which implies a blocked road. Although all of them capture this state of jumping, DAAGCN's pattern-matching ability is more robust due to its prediction curve is closer to the trend of the actual value.

\begin{figure}[!t]
\centering
\subfloat{\includegraphics[width=4.0in]{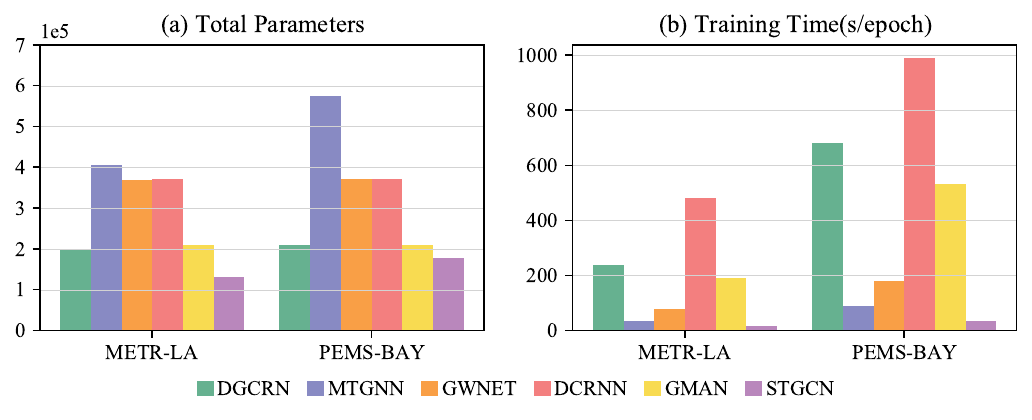}%
\label{b1}}
\hfill
\subfloat{\includegraphics[width=4.0in]{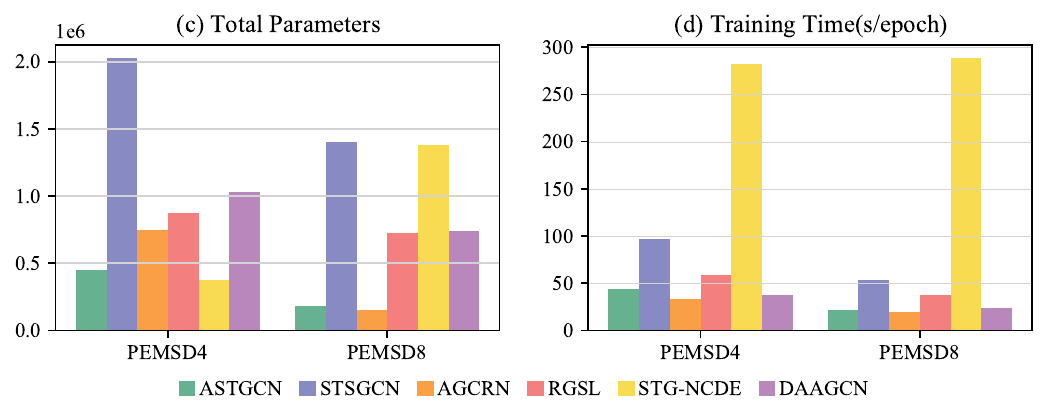}%
\label{b2}}
\caption{Summary of parameters and efficiency of different models in traffic datasets.}
\label{fig:summary}
\end{figure}

The goal of traffic forecasting is to pursue models with low overhead, low complexity, and high generalization capability, which directly relates to the offline deployment of the models. Therefore, it is necessary for us to systematically study the system resource consumption and computational efficiency of STGNNs. Fig. \ref{fig:summary} shows the number of parameters and training time per epoch for different models on different datasets. Generally speaking, the larger the dataset, the more system resources and training time the model will consume. On the same dataset, we found that the training time of the model is not strictly correlated with the number of parameters, but is more related to the model architecture and operators. For example, on METR-LA and PEMS-BAY, GWNET and DCRNN have similar parameter counts, but the training efficiency of GWNET is 5 times that of DCRNN. DGCRN and DCRNN use the same Seq2Seq architecture, but after DGCRN introduced a curriculum learning strategy, it effectively improved the model's training efficiency. DGCRN believes that there is no need to calculate the error of all time steps in the early stage of training, but rather to gradually increase the model's prediction length as the number of iterations increases, guiding the model training in a gradual convergence manner.

The number of parameters is closely related to the specific hyperparameters of the model, and it reflects the complexity of the model to a certain extent, affecting the utilization of GPU memory. For example, STSGCN constructs a local spatio-temporal graph and performs multi-level cascading, which dramatically increases the number of parameters, and the corresponding training time also increases significantly. In general, graph learning methods (such as AGCRN, RGSL, DAAGCN, etc.) have a higher dependence on parameters, mainly reflected in the influence of node embedding dimensions on spatial representation learning, especially when dealing with datasets with a large number of nodes. Furthermore, we can find that although STG-NCDE achieved good prediction performance, the neural control differential operators used generated huge intermediate parameters, which significantly increased the GPU memory occupation and computation time. MTGNN and GWNET have similar performance and the same model architecture, but after MTGNN optimized the temporal convolutional module and graph learning layer in the structure, the training time was effectively shortened. Therefore, good model design can not only reduce computational complexity but also improve the model's generalization ability. If we continue to increase the complexity of the model in exchange for performance improvements, it will exacerbate the dilemma of spatio-temporal graph model design.

\subsection{Usability And Practicality}
With the growing number of spatial-temporal traffic prediction models, we develop a training pipeline called STG4Traffic to provide a convenient and standardized, scalable project architecture for subsequent research. It is organized as presented in Fig. \ref{fig:pipline} below.

\begin{figure}[!h]
\centering
\includegraphics[width=0.90\linewidth]{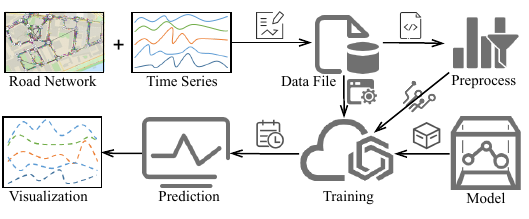}
\caption{The Pipeline of Benchmark for Traffic Prediction.}
\label{fig:pipline}
\end{figure}

STG4Traffic mainly interacts with users through a unified configuration file, which contains two sub-tasks for traffic speed and traffic flow prediction. These two subtasks provide a common standardized data interface and method interface for model design: DATA directory stores raw traffic data resources and pre-processed data files; LIB is a toolkit that designs callable data loading methods, evaluation methods, and some basic methods; LOG stores the project run logs and saves the final training model; MODEL is a model design file, which realizes the decoupling of model, data and training process. Researchers can extend the data and perform self-defined model designs according to their demands. Taking the DCRNN design as an example, we first complete the model definition in MODEL and then create the DCRNN directory in the subtask directory, where we create: DCRNN\_Config.py, DCRNN\_Utils.py [optional], DCRNN\_Trainer.py, DCRNN\_Main.py, and DATANAME\_DCRNN.conf [multiple]. The functions and meanings of these files are listed below:

\begin{enumerate}
    \item DATANAME\_DCRNN.conf is the configuration of model parameters and experimental setup, which can vary based on the dataset being used.
    \item DCRNN\_Config.py is responsible for reading the preset configuration file.
    \item DCRNN\_Utils.py defines additional methods that are not available in the public interface.
    \item DCRNN\_Trainer.py serves as the trainer for the model, handling the training, validation, and testing processes.
    \item DCRNN\_Main.py acts as the entry point for the project, managing tasks such as data loading, model and parameter setting, and more.
\end{enumerate}

In the initial phase of benchmarking, we pre-selected 18 high-impact works in the field on two types of transportation data to serve as the basis for the design of our framework. These case studies provide excellent learning examples for beginners of spatial and temporal mapping tasks to better understand the concepts and techniques. In addition, mature developers can quickly implement their mature research ideas by calling on pre-implemented interfaces using the existing baseline assessment results of the benchmark. The source code of this open-source project is available at https://github.com/trainingl/STG4Traffic.

This training pipeline based on configuration files simplifies the code design of model training, provides a flexible parameter configuration interface to fine-tune existing models, and has good scalability. For users, as long as they understand the functions of each file in the benchmark project in advance, and call the standard input and output interfaces designed by STG4Traffic, they can ignore the setup of data processing and training process, and focus on model design to achieve rapid iteration and effective method innovation.

\section{Discussions on Future Directions}
Over the years, significant advancements have been achieved in traffic forecasting using STGNNs. However, there are still several challenges that require further attention and resolution. 
In this section, we highlight some future research directions for addressing the challenges.

\textbf{Data Quality}: Data collected from sensors often suffer from noise or contain a certain percentage of missing data (zero-value padding during preprocessing). While we address this issue carefully during test evaluation, the training process is highly susceptible to being influenced by these anomaly samples. Additionally, the narrow time frame of data sampling, represented by parameters $P=Q=12$, poses challenges in capturing the periodicity of long sequences on one hand, and makes learning similarity graphs of node pairs unreliable on the other \cite{DBLP:conf/kdd/ShaoZWX22}. Moreover, the datasets collected in the study are limited, lacking the introduction of enriched external information or metadata such as weather, events, dates, etc. \cite{DBLP:journals/isci/BaoLSCDS23} and the fusion of these heterogeneous data presents a persisting challenge. 
Some studies have explored effective approaches to addressing data quality concerns in traffic forecasting. For instance, graph contrastive learning methods \cite{DBLP:conf/www/0001XYLWW21, DBLP:journals/corr/abs-2301-12603} have shown promise in combating training noise and capturing robust spatial-temporal features. Time Series Pre-training techniques \cite{DBLP:conf/kdd/ShaoZWX22} have also been found effective for long time series data with periodicity. In the ST-MetaNet model \cite{DBLP:conf/kdd/PanLW00Z19}, meta-knowledge is leveraged to parameterize the model, while considering heterogeneous information in the road network to model spatial-temporal dependencies and achieve improved feedback. These studies provide valuable insights and serve as strong references for addressing concerns related to data quality.

\textbf{Research  Perspectives}: Research on traffic forecasting mainly revolves around the modeling of temporal and spatial correlations. (1) \textbf{Temporal Heterogeneity}: Existing approaches that model temporal correlation for all nodes often adopt a shared parameter space, ignoring the temporal heterogeneity among different nodes. STID \cite{DBLP:conf/cikm/ShaoZ00X22} argues that this difference in temporal patterns is due to spatial heterogeneity, and it identifies this indistinguishability in combination with multiple spatial-temporal embeddings. Spatial-Temporal Self-Supervised Learning (ST-SSL) \cite{DBLP:journals/corr/abs-2212-04475} proposes a novel paradigm that enhances the representation of traffic patterns to capture both spatial and temporal heterogeneity by augmenting traffic-level and topology-level data. (2) \textbf{Dynamic Graph}: Although adaptive graphs can compensate for the knowledge bias caused by pre-defined graphs, they can only represent fixed node relations after training and cannot be dynamically adjusted with the data characteristics. Some work as DGCRN \cite{DBLP:journals/corr/abs-2104-14917} and D$^2$STGCN \cite{DBLP:journals/pvldb/ShaoZWWXCJ22}, among others, take into account the basic fact that the location dependence of a road network changes dynamically with time. They propose a time-driven dynamic connectivity graph that infers instantaneous connection patterns based on the current traffic state. Although some related work exists, the design of dynamic graphs remains a significant technical challenge. (3) \textbf{Long-Range Dependence}: Attention mechanisms are very flexible in global spatial-temporal modeling, especially in capturing long-range temporal dependencies with good performance, and time series models based on Transformer variants are widely used in various fields. Sequence pattern decomposition \cite{wang2023micn} is also a potential solution for the future. Long-term dependence information is inferred from three structural components of time series trends, periods and residuals to improve prediction performance. (4) \textbf{GNN-Free Method}: MLP-based spatial-temporal prediction models \cite{DBLP:conf/cikm/ShaoZ00X22, DBLP:conf/kdd/EkambaramJNSK23} have become a new idea of spatial-temporal prediction with its advantage of balancing efficiency and performance. It does not advocate the use of complex STGNN modeling spatial-temporal dependence, but adopts diversified spatial-temporal information encoding, and combines the most simple MLP model to process spatial-temporal data. However, essentially, the MLP-Based approaches also draw on graph signal propagation or diffusion to represent neighborhood node correlation, and this viewpoint revisits the features behind spatial-temporal and pays more attention to the research and design of the Embedding layer.

\textbf{Migration Methods}: In the realm of STGNNs, the integration of various methods from different fields has shown the potential to enhance model performance. Generative adversarial networks \cite{DBLP:journals/tits/ZhangWCCH21, DBLP:journals/corr/abs-2208-03063} combat training generators (STGNNs) and discriminators by comparing the proximity of the predicted information to make the predictions converge to the ground truth as much as possible. Knowledge graphs \cite{DBLP:conf/jist/YangQ22, DBLP:journals/tits/ZhuHDTZWLL22, DBLP:conf/icpr/MallickBRM20} have also emerged as a valuable tool, establishing relationships among different transportation entities and concepts. They enable the consolidation and sharing of knowledge from diverse fields, facilitating comprehensive analysis and prediction of the entire traffic system. Recent studies such as AutoSTG \cite{www/PanKYLYZZ21} and AutoCTS \cite{DBLP:journals/pvldb/WuZGHYJ21} propose the use of automated machine learning to streamline the construction and optimization of traffic prediction models. These approaches efficiently build STGNNs that meet performance requirements while emphasizing accuracy.  Additionally, transfer learning \cite{DBLP:conf/itsc/HuangSZY21, DBLP:conf/icpr/MallickBRM20} has demonstrated its usefulness by leveraging models trained in other cities or regions with similar traffic patterns to initialize new models. This approach accelerates the training process and alleviates challenges arising from limited data availability. The advancement of these techniques has diversified the research approaches for traffic forecasting, providing greater potential for solving the problems in this domain. However, the transferred methods from other domains have not been well applied due to the differences in data distribution and domain knowledge, which poses challenges to their effective utilization.

\section{Conclusion}
In this paper, we first present a systematic survey of graph design and graph computation techniques for traffic prediction. Then we provide a detailed introduction to the key modeling components, technical details, and well-established methods of spatial-temporal graph neural networks. In order to establish a standardized benchmark, we introduce STG4Traffic, providing a thorough overview of the performance and efficiency of various methods within this framework. Finally, we conduct an analysis of the challenges encountered in this study and summarize potential solutions for future investigations. We hope that this research make a positive and impactful contribution to the field of spatial-temporal prediction.

\section{Acknowledge}
Chunjiang Zhu is supported by UNCG Start-up Funds and Faculty First Award. Detian Zhang is partially supported by the Collaborative Innovation Center of Novel Software Technology and Industrialization, the Priority Academic Program Development of Jiangsu Higher Education Institutions.




\bibliographystyle{elsarticle-num} 
\bibliography{elsarticle}
\end{document}